**IEEE** *Access*

Multidisciplinary | Rapid Review | Open Access Journal

# RESEARCH ARTICLE

# Generating Synthetic Malware Samples Using Generative AI

**TIFFANY BAO**[ID]1, **KYLIE TROUSIL**[ID]2, **QUANG DUY TRAN**[ID]3,
**FABIO DI TROIA**[ID]3,4, **(Member, IEEE)**, AND **YOUNGHEE PARK**[ID]4,5, **(Member, IEEE)**
[1]Department of Computer Science, Boston University, Boston, MA 02215, USA
[2]Department of Computer Science, University of Wisconsin–La Crosse, La Crosse, WI 54601, USA
[3]Department of Computer Science, San José State University, San Jose, CA 95192, USA
[4]Silicon Valley Cybersecurity Institute, San Jose, CA 95192, USA
[5]Department of Computer Engineering, San José State University, San Jose, CA 95192, USA

Corresponding authors: Younghee Park (younghee.park@sjsu.edu) and Fabio Di Troia (fabio.ditroia@sjsu.edu)

This work was supported by the National Science Foundation (NSF) under Award #2244597.

**ABSTRACT** Malware attacks have a significant negative impact on organizations of varied scales in the field of cybersecurity. Recently, malware researchers have increasingly turned to machine learning techniques to combat sophisticated obfuscation methods used in malware. However, collecting a diverse set of malware samples with various obfuscation techniques is challenging and often takes years, especially for newly developed malware. This issue is further compounded by a well-known limitation of machine learning models: their poor performance when training data is scarce. In this paper, we propose a new system for generating synthetic malware samples to augment imbalanced malware dataset. Our approach decomposes malware binary samples into mnemonic opcode sequences, leveraging natural language processing to extract contextual meaning behind malware opcode features to aid the learning of generative AI (GenAI) employed in this paper, Generative Adversarial Networks (GAN), Wasserstein Generative Adversarial Networks with Gradient Penalty (WGAN-GP), and a modified Diffusion model. The experiment results show that augmenting training data with Diffusion-based synthetic data significantly improves classification performance for minor classes by up to 60% on average. This enhancement ultimately leads to an overall malware classification performance of 96%, an 8% improvement. These findings demonstrate the high quality and fidelity of the synthetic data, its robustness, and its potential applications in malware analysis. Specifically, synthetic malware data proves effective in improving the classification of minor malware classes and detection rates, even though the size of known malware data is significantly small.

## I. INTRODUCTION

Malware is malicious software intentionally designed to disrupt, damage, or gain unauthorized access to computer systems [1], [2]. With the increasing sophistication of malware, adversaries often employ obfuscation techniques to hide malicious behavior, such as opcode obfuscation [3]. According to SonicWall, the number of malware attacks rose by 10% in 2023, reaching 6.06 billion attacks globally [4]. These attacks can lead to severe consequences like data loss, reduced system performance, and compromised access to files.

Traditional malware detection methods have been widely studied over the past decade due to their effectiveness in identifying known threats and adapting to evolving attack patterns [5]. Both traditional machine learning and deep learning techniques, including Random Forest, MLP, CNN, and RNN, have been commonly used for malware classification [2], [6], [7]. Several studies have also applied NLP techniques—such as Word2Vec, HMM2Vec, ELMo, and BERT—by converting binary malware into mnemonic opcode sequences for analysis [7], [8], [9], [10]. Harshit et al. [9] proposed using HMM and modified GAN models to generate synthetic malware for augmenting training datasets. Similarly, Tran and Di Troia [8] explored various NLP techniques to enhance the learning capabilities of WGAN-GP.

Despite the effectiveness of many machine learning techniques in malware classification, challenges remain when dealing with limited datasets. Malware patterns frequently change due to varying conditions and environments. Additionally, attackers often obfuscate their behavior to evade detection, making it difficult to extract reliable features. This, combined with the continuous emergence of new malware, results in insufficient training data, preventing machine learning models from achieving optimal performance. To address this challenge, developing a generative system to create high-quality malware samples can help augment training data and improve classification accuracy.

This paper proposes a novel system for generating synthetic malware samples to address imbalanced malware datasets. The system consists of four main components: (1) collecting and decomposing malware data into mnemonic opcode sequences, (2) extracting malware features using Natural Language Processing (NLP) techniques, (3) training generative models to create synthetic data, and (4) evaluating the quality, robustness, and applicability of the generated samples. By generating synthetic malware, the system effectively enhances imbalanced training datasets, improving classification performance for minor malware classes, which have a very small number of samples for training. Extensive evaluations demonstrate that the system achieves high performance and robustness, with the potential to mitigate zero-day attacks.

Our system introduces several key contributions:

1) This paper contributes to the development of a synthetic malware generator designed to produce high-quality synthetic data, enhancing the performance of malware classification
2) Our system generated high-quality synthetic data, effectively augmenting the limited training data
3) This paper contributes to the modifications to generative model architectures, enabling better adaptation to opcode sequence data
4) Our system introduces new, comprehensive performance metrics to evaluate the quality, robustness, and applicability of synthetic data.

The rest of the paper is structured as follows: Section II reviews related work and provides background on our generative models. Section III details the proposed system architecture. Section IV describes the implementation and experimental results. Section V examines the strengths and limitations of our approach. Finally, Section VI summarizes the findings and outlines directions for future work.

## II. BACKGROUND
### A. RELATED WORK
Malware classification is gaining more and more awareness due to the fast growth and evolution of malware in current days. Accurately detecting and classifying malware data for defense purposes is one of the most important tasks in the cybersecurity field. To achieve this goal, many malware detection techniques have been proposed by using different deep-learning networks. Deep learning models, such as Multilayer Perceptron (MLP) [11], Convolutional Neural Networks (CNNs) [12], and Recurrent Neural Networks (RNNs) [13], have been employed to detect both static and dynamic malware behaviors with impressive accuracy. MLP, as a feedforward neural network, has been widely adopted for classifying malware by analyzing static features like opcode sequences and dynamic features such as API calls [14] and network activities. While MLP is effective for non-sequential data, more advanced models like CNNs and RNNs offer greater flexibility for handling structured or sequential data. CNNs, in particular, excel at transforming malware binaries into image-like representations, enabling the detection of intricate patterns within the data [6]. Based on previous research [2], these deep learning models have significantly improved the precision and robustness of modern malware detection systems.

What's more, new approaches proposed to use Natural Language Processing techniques have been increasingly applied to malware detection, particularly for feature extraction and classification tasks. Approaches utilizing models such as BERT, Word2Vec, HMM2Vec, and ELMo [7] have demonstrated their ability to convert sequential malware data, such as opcode sequences and API calls, into meaningful vector embeddings. These embeddings act as high-quality inputs for machine learning classifiers, allowing for improved accuracy in distinguishing between malware families. Similar studies [15], [16], [17] have also demonstrated the effectiveness of NLP techniques for decomposing malware at the pseudo-code level. Researchers found that NLP-based word embeddings enhance the feature representation and allow models to learn complex behaviors that are difficult to detect using traditional methods.

Despite advancements in malware detection techniques, one recurring challenge is the limited availability of labeled malware samples for training models. This shortage of data can restrict the ability of machine learning models to generalize and effectively classify diverse malware types. To address this issue, researchers have explored various methods for generating synthetic malware, which can be used to augment training datasets. GAN [18] and VAE [19] models have been used to generate 2-dimensional images and have been proven useful for malware generation. However, converting malware files to images is computationally expensive. Additionally, the training and testing process can be inevitably time-consuming. Therefore, instead of using 2-dimensional images, another approach is to use opcode sequences, which largely decreases the computation time. Harshit and Q. D. works [8], [9] both used opcode sequences to represent malware and used different NLP techniques, such as BERT [20], and experimented with different generative models such as GAN and WGAN-GP. Such an approach gives good results and takes comparably less time. This demonstrates the significant reduction in time needed to train the models and generate realistic synthetic samples.

In this paper, we extend previous research by employing GAN, WGAN-GP, and the Diffusion model, a non-adversarial generative network, to generate synthetic opcode sequences for malware. For feature extraction, we apply Word2Vec [21], transforming the opcode sequences into one-dimensional datasets. These embeddings are then used as input for traditional classifiers, including MLP [11], Random Forest [22], and SVM [23], to assess the effectiveness of our generated samples in malware classification. Our primary focus is on comparing the performance of the three generative models (GAN, WGAN-GP, and Diffusion) in producing realistic synthetic malware samples. By evaluating these models, we aim to provide insights into their relative advantages in generating high-quality data for improving malware classification accuracy and reducing training times.

## III. METHODOLOGY

### A. SYSTEM ARCHITECTURE

Figure 1 presents the system architecture diagram for our research. The diagram outlines each component and sub-component of our system. The workflow begins with collecting and cleaning malware data from various infected Windows operating systems as part of the data preprocessing step. The cleaned data is then stored back into the database and used to extract tokenization and embeddings, which capture the characteristics, signatures, and patterns of each malware family. These embedding vectors are then used as inputs to train generative models, including GAN, WGAN-GP, and Diffusion. Each model's training process is distinctive due to its specific architecture. Throughout this research, these embeddings are referred to as "real" embedding samples, as they are directly derived from the input dataset. After completing the training phase, all generative models produce synthetic malware embeddings as outputs. We then evaluate the quality of these outputs using various metrics: binary classification, visualization through dimension reduction techniques, synthetic-only training evaluation, malware classification, fidelity score, and similarity score. Each metric provides a distinct perspective on the similarity between real and generated embeddings.

### B. PREPROCESSING

The dataset comprises approximately 40,000 samples from 25 different malware families, sourced from two large public datasets: Malicia [24] and VirusShare [25]. We collected malware data from these sources to ensure diversity in malware behaviors, enhancing the generalizability and robustness of our findings. We select the top three families from Malicia, each containing over 1,000 samples: Zbot, Zeroaccess, and Winwebsec. The remaining 22 families were sourced from VirusShare [25], where we selected the largest available families, excluding Zbot and Winwebsec to avoid duplication with the Malicia repository. The family sizes and distribution details are shown in Table 2.

We then extracted assembly opcode sequences from the executable files in the Malicia and VirusShare repositories, excluding any corrupt files. An example of the extracted opcode sequences is shown in Table 1.

**TABLE 1.** Opcode sequence examples.

| Malware Sample # | Number of Opcodes | Opcode Sequence |
|---|---|---|
| $s_0$ | 3 | [mov, sub, call] |
| $s_1$ | 245 | [xor, mov, push, ..., pop, jmp, add] |
| $s_2$ | 578 | [add, mov, cmp, ..., jmp, sub] |
| $s_3$ | 780 | [mov, cmp, ..., push, pop, jmp, call] |
| $s_4$ | 2044 | [lea, xor, mov, ..., push, call, jmp] |

During preprocessing, we identified duplicate samples across families. These duplicates were retained in the training set to increase dataset size, enhancing model learning. However, for testing, all duplicates were removed to create a "unique" dataset, ensuring fair evaluation and preventing artificially inflated performance metrics.

**TABLE 2.** Dataset summary.

| Malware Family | # of Samples | # of Unique Samples | Source |
|---|---|---|---|
| Lamechi.B | 971 | 2 | VirusShare |
| Diplugem | 2269 | 3 | VirusShare |
| Systex.A | 1098 | 4 | VirusShare |
| Startpage | 1313 | 170 | VirusShare |
| Delf | 1679 | 197 | VirusShare |
| Enterak.A | 1530 | 365 | VirusShare |
| DelfInject | 839 | 425 | VirusShare |
| Zbot | 2136 | 484 | Malicia |
| Small | 1051 | 533 | VirusShare |
| CeeInject | 903 | 643 | VirusShare |
| Renos | 864 | 652 | VirusShare |
| Toga!rfn | 985 | 843 | VirusShare |
| Hotbar | 844 | 840 | VirusShare |
| Injector | 1161 | 927 | VirusShare |
| FakeRean | 1089 | 1065 | VirusShare |
| Expiro.BK | 1095 | 1095 | VirusShare |
| Zeroaccess | 1305 | 1161 | Malicia |
| OnLineGames | 1366 | 1174 | VirusShare |
| Allaple.A | 1294 | 1294 | VirusShare |
| VBInject | 1688 | 1545 | VirusShare |
| Beebone | 1629 | 1568 | VirusShare |
| Winwebsec | 4355 | 1586 | Malicia |
| Obfuscator | 2102 | 1735 | VirusShare |
| Vundo | 1877 | 1792 | VirusShare |
| Vobfus | 4204 | 4179 | VirusShare |

### C. NATURAL LANGUAGE PROCESSING

Given the cleaned opcode sequences as input data, we tokenized each opcode in the dataset. Next, using the Word2Vec technique, we converted the tokenized opcodes into numerical vectors, with each vector representing an opcode. At this stage, each initial malware file is represented as many numerical vectors. Due to opcode obfuscation in metamorphic malware, the number of opcodes per malware file varies. Therefore, to create a malware embedding, we averaged all the opcode vectors in a file, producing a single embedding for each malware file. This process is repeated for all malware files within each family and is visualized through the pipeline in Figure 2 with malware sample $s_0$ as an example.

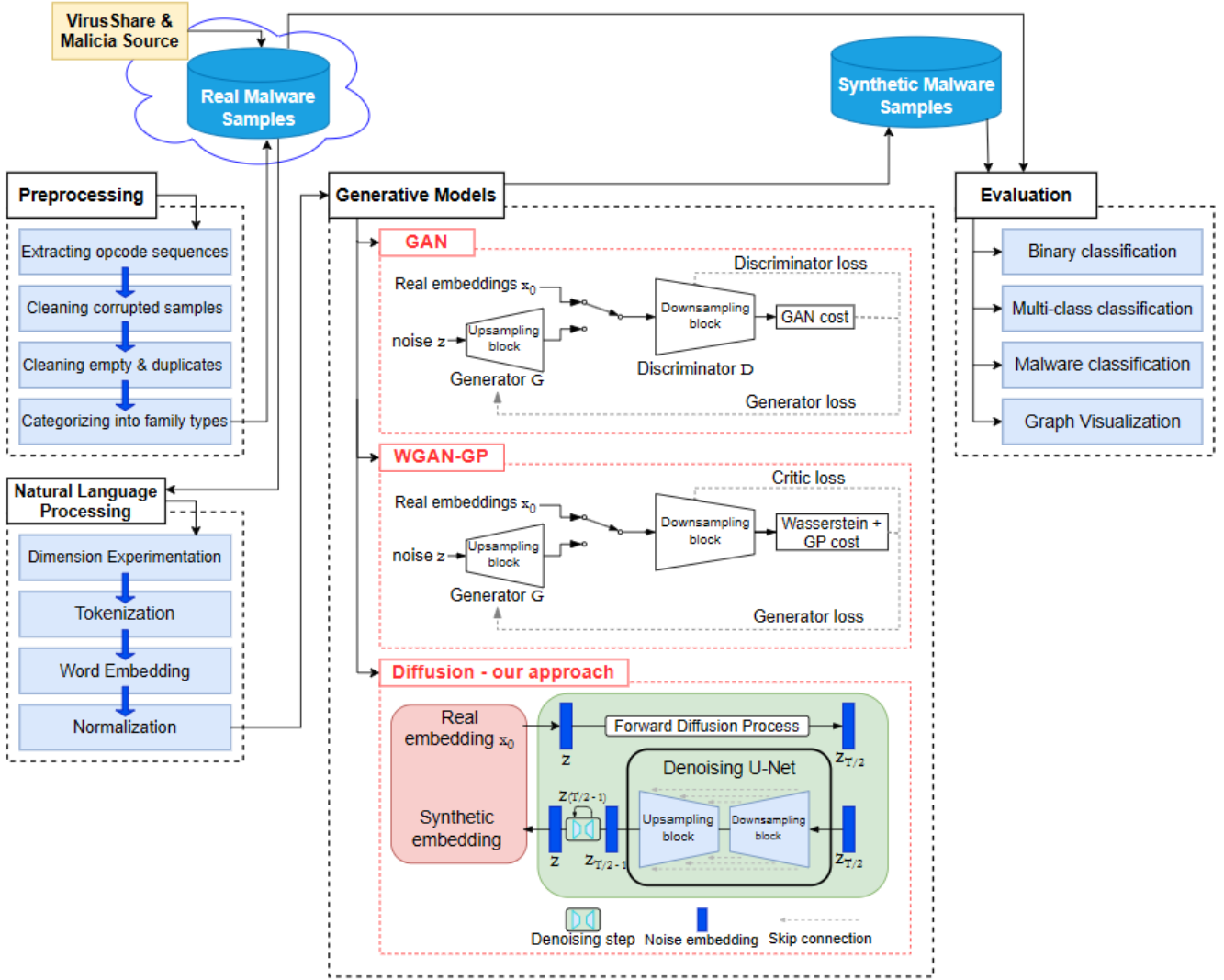

**FIGURE 1.** System architecture.

Note that the choice of NLP technique is not central to our study, as our primary focus lies in evaluating the performance of generative models. Therefore, the specific NLP approach used here is secondary to our main objective.

### D. GENERATOR
We take the embedding malware as input to train each of the generative models: GAN, WGAN-GP, and Diffusion. These three models were selected for specific purposes: GAN serves as the baseline generative model, WGAN-GP is included as it represents the state-of-the-art approach for malware generation according to previous research, and the Diffusion model is explored as a novel approach, as it has not yet been applied to malware domain in the literature.

The key difference between the Diffusion model and GANs lies in their architectures: GANs are adversarial networks, while Diffusion is a non-adversarial process. Due to this structural difference, we decided to include the Diffusion model to evaluate whether it can offer superior performance in generating malware opcode sequences.

#### 1) GENERATIVE ADVERSARIAL NETWORKS
The GAN architecture in our research is illustrated in Figure 3, following the similar structure as the traditional GAN introduced by [26], it consists of two neural networks, a generator $G$ and a discriminator $D$, trained in opposition to one another. However, unlike traditional GAN architectures, which are primarily designed for generating 2D images using convolutional layers, our GAN model is adapted to generate one-dimensional sequential malware embeddings.

In Generator G, the input is a random noise numerical vector $z$, which is processed through multiple dense layers.

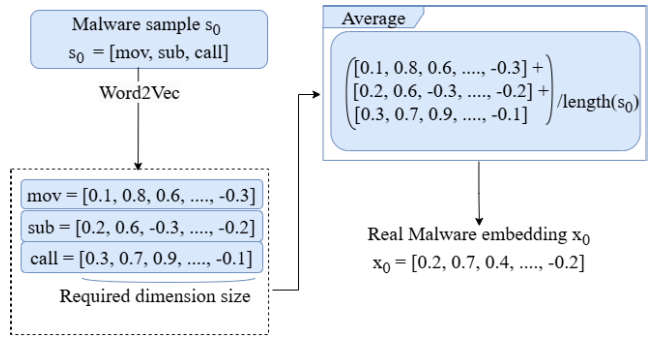

**FIGURE 2.** Pipeline for generating malware embeddings from a sample malware file $s_0$. The Word2Vec model converts each opcode into a 104-dimensional vector, shown in two scaling ranges, $[-1, 1]$ and $[0, 1]$. These vectors are then averaged to produce a single malware embedding for each file.

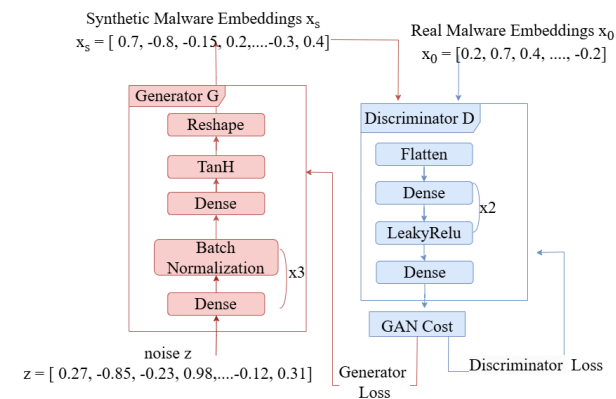

**FIGURE 3.** Modified GAN architecture.

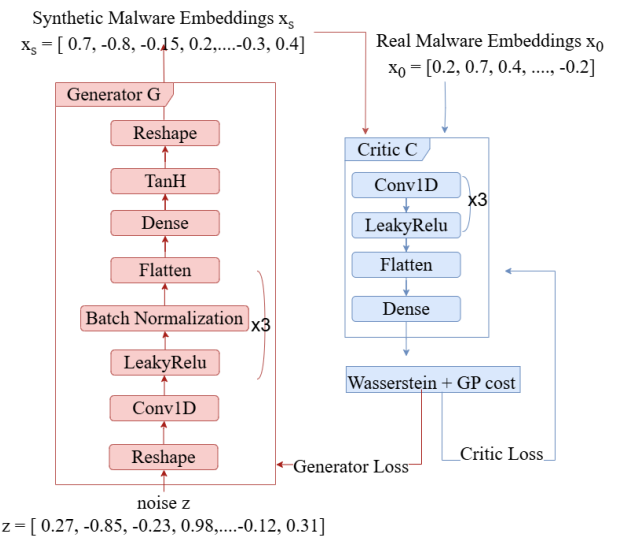

**FIGURE 4.** Modified WGAN-GP architecture.

These layers transform $z$ into synthetic malware embeddings $x_s$, capturing the essential patterns in the real malware data. Each layer is followed by batch normalization and a Tanh activation function, which help stabilize training and improve output quality.

The discriminator $D$ receives real malware embeddings $x_0$, obtained from the NLP preprocessing stage, and synthetic embeddings $x_s$ from the generator as inputs. Its task is to classify each embedding as either real or synthetic. By comparing $x_0$ with $x_s$, the discriminator learns to distinguish between genuine and generated embeddings and then helps the generator to produce synthetic malware embeddings that increasingly resemble the real embeddings.

### 2) WASSERSTEIN GENERATIVE ADVERSARIAL NETWORKS WITH GRADIENT PENALTY

The WGAN-GP architecture in our approach comprises a generator $G$ and a critic $C$, as illustrated in Figure 4. It shares a similar structure with the original WGAN-GP setup introduced by Gulrajani et al. [27]; it uses a gradient penalty to ensure a reliable enforcement of the Lipschitz constraint and results in enhanced model performance. However, instead of using Conv2D, we modified our WGAN-GP model to use Conv1D and Dense Layers. These layers are specifically designed to capture the sequential nature of opcode data, with

Conv1D layers enabling the model to learn patterns within one-dimensional sequences.

In the generator $G$, it begins with a numerical noise vector $z$, which is transformed into synthetic malware embeddings $x_s$ through a series of Conv1D and dense layers, and outputs to the Critic $C$.

The critic $C$ receives both real malware embeddings $x_0$ from the NLP preprocessing and synthetic embeddings $x_s$ generated by $G$. It distinguishes real embeddings from synthetic ones while maintaining the gradient penalty to ensure stability in training. The penalty enforces smooth gradients in the critic's output, this helps the generator to produce synthetic malware embeddings with higher qualities.

### 3) DIFFUSION

Denoising diffusion probabilistic models, inspired by nonequilibrium thermodynamics, represent an innovative technique for generative models, introduced by the authors in [28]. The model begins with the diffusion process, taking the embedded malware samples as input and gradually adding Gaussian noise using a Markov chain to create a noisy, destroyed sample as output. This forward diffusion process of adding noise has no learnable parameters and is a constant during training.

To create synthetic malware embeddings, the model then tries to reconstruct malware embeddings from the noisy samples produced by forward diffusion. This process requires estimating the noise added at each step, as the exact value is unknown. Therefore, a U-Net architecture is employed to predict the amount of noise added at each step, allowing the model to iteratively remove noise at each timestep until a synthetic sample is produced. Through repeated training, the model can create samples analogous to the original, eventually allowing the model to generate synthetic malware embeddings from an initial sample of Gaussian noise. This process is visualized through Figure 5, where each box corresponds to the sample embedding with the size denoted on the side of the box. White boxes represent the copied samples, and the arrows denote different operations.

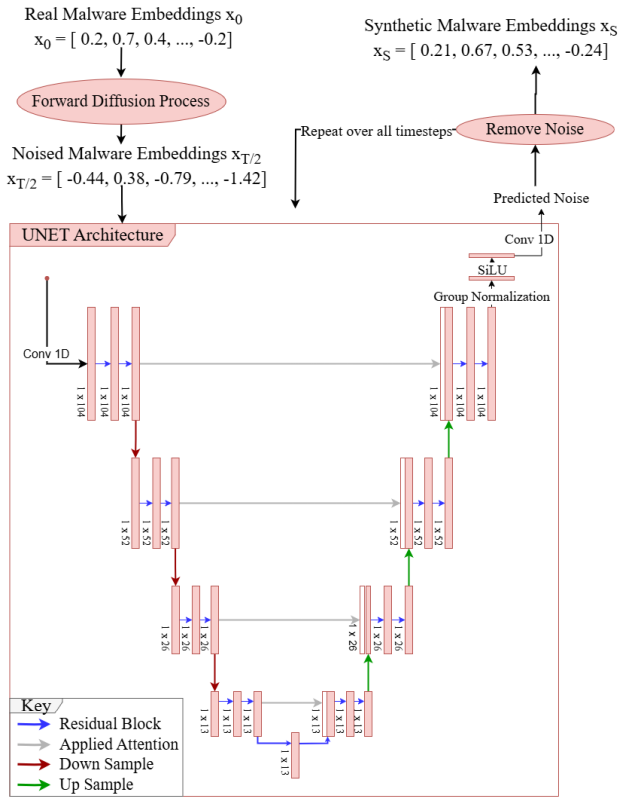

**FIGURE 5.** Modified diffusion architecture.

### E. EVALUATION SYSTEM

Figure 1 proposes six key metrics to validate the quality of our generated data. Together, these metrics ensure the data's quality and the robustness of the model while highlighting the impact of the synthetic samples.

The first metric, binary classification, measures how realistic the synthetic samples are compared to real data.

The second metric, dimension reduction techniques, specifically t-Distributed Stochastic Neighbor Embedding (t-SNE) [29], are employed to visualize the samples, providing insights into the sample similarity.

The third metric is Synthetic-Only Training Evaluation, which is used to examine the synthetic samples' distribution and alignment. By training classifiers exclusively on synthetic data and testing on real data, we replicate the principles of the Fréchet Inception Distance (FID) [30].

Next, Threshold-Based Malware Classification was used to evaluate the effectiveness of synthetic data augmentation and value the implications of synthetic malware data. By introducing a thresholding mechanism, this method determines the optimal augmentation strategy, especially for small or underperforming malware families, highlighting the trade-offs between dataset size and classification accuracy.

In the next metric, we assess the fidelity by training on real data and testing on synthetic samples. This metric provides a comprehensive framework for evaluating the quality and utility of the generated synthetic malware.

The final metric demonstrates the robustness of our model through the cosine similarity score [31]. The goal of this metric is to achieve a similarity score close to the original, proving the sample distributions are similar.

## IV. EVALUATION
### A. IMPLEMENTATION
#### 1) PREPROCESSING
To disassemble the binary files into assembly code, we utilized objdump, a command-line program part of the GNU Binary Utilities library for Unix-like operating systems. Then, the assembly code was processed using a Python script to remove information such as registers, labels, and addresses, leaving only the opcodes. These opcode files are then input into the NLP model.

Objdump command: `objdump -d $filename | sed "/[^\t]*\t^\t]*\t/!d" | cut -f 3 | sed "s/.*$//" > "$output_path"/basename "$filename".txt`.

The `objdump -d $filename` disassembles the binary file specified by `$filename` and produces assembly code, where each line typically contains the instruction address, bytes, and the corresponding opcode. Next, `sed "/[^\t]*\t^\t]*\t/!d"` retains only lines that match the expected structure (address, bytes, opcode), discarding any lines that do not contain an opcode. Then, `cut -f 3` command extracts the third field in each line, isolating the opcode. Finally, `sed "s/.*$//"` removes any additional information following the opcode, ensuring that only the opcode itself remains. The output is saved in a text file `"$output_path"/basename "$filename".txt`.

#### 2) NATURAL LANGUAGE PROCESSING
We applied the Word2Vec method in our natural language processing approach. In our process, we treated the opcodes within each malware family as separate corpora and created Word2Vec embeddings for each malware file. Specifically, for each family, we trained a distinct Word2Vec model and generated a vector for each file by averaging the embedding vectors of the opcodes. We refer to these embedding vectors as the "real embeddings."

To determine the optimal embedding dimensionality for effective malware classification, we performed multiclass classification on the real embeddings of different dimensions (64, 104, and 128 dimensions) and then evaluated the F1 scores. Higher F1 scores indicate better performance in classifying malware across different families. As shown in Table 3, the dimension of 104 achieved the highest F1 score (0.932), leading us to select this dimensionality for all generative models.

For all three generative models, we converted each opcode into a 104-dimensional vector using Word2Vec and averaged these vectors within each malware file to create a single malware embedding. Then, we scaled these 104-dimensional embeddings to the range $[-1, 1]$. This matches the requirements of the Tanh activation function used in the GAN and WGAN-GP networks. Furthermore, this scaling ensures compatibility with previous models and leads to stable training because of the symmetry of the data centered around zero.

Table 4 provides an overview of the selected vector size and normalization range for each model.

**TABLE 3.** Embedding dimension evaluation.

| Dimension | F1 |
|-----------|-------|
| 64 | 0.928 |
| 104 | 0.932 |
| 128 | 0.915 |

**TABLE 4.** Baseline model evaluation.

| Model | Vector Size | Normalization Range | F1 Score |
|-------|-------------|---------------------|----------|
| Diffusion | | | |
| WGAN-GP | 104 | [-1,1] | 0.87 |
| GAN | | | |

### 3) GENERATOR

In this subsection, we highlight the modifications made to adapt the GAN, WGAN-GP, and Diffusion models to fit our malware dataset, which involves generating sequential 1D data, where each sequence represents a series of opcodes.

In the GAN model, we replaced traditional convolutional layers with fully connected dense layers to handle the one-dimensional vector of the malware embeddings. The generator starts with a one-dimensional noise vector, which is processed through multiple dense layers with batch normalization and Tanh activation functions. This processed noise vector is transformed into a synthetic malware embedding, output as a numerical vector that approximates the characteristics of real malware embeddings. This synthetic embedding is then fed into the discriminator. The discriminator, also composed of dense layers, processes each embedding and assigns a probability score indicating whether the embedding is real or synthetic. This setup enables the GAN to produce synthetic malware embeddings that closely resemble real embeddings by capturing essential patterns within the one-dimensional data.

In the WGAN-GP model, we implemented a structure using 1D convolutional layers (Conv1D). The generator network includes three Conv1D layers followed by a fully connected layer, with Tanh activation applied at the output layer. Starting from a one-dimensional noise vector, the generator produces a synthetic malware embedding, which is then passed to the critic. The critic, also consisting of three Conv1D layers, uses LeakyReLU activation functions to improve gradient flow, and a gradient penalty is applied to enforce the Lipschitz constraint, stabilizing training. Table 5 summarizes the specific parameters and configurations for both the GAN and WGAN-GP models.

**TABLE 5.** GAN and WGAN-GP hyperparameters.

| | Parameter | GAN Value | WGANGP Value |
|---|-----------|-----------|--------------|
| **Generator** | Activation | Tanh | |
| | Dense_layer_1 / Conv1D_layer_filter_1 | 200 | 64 |
| | Dense_layer_2 / Conv1D_layer_filter_2 | 400 | 32 |
| | Dense_layer_3 / Conv1D_layer_filter_3 | 800 | 16 |
| | BatchNorm_momentum | 0.8 | 0.8 |
| | kernel_size | N/A | 3 |
| | Dense_output / Conv1D_padding | 100 | same |
| **Discriminator / Critic** | Activation | Sigmoid | LeakyReLU |
| | Dense_layer_1 / Conv1D_layer_filter_1 | 400 | 64 |
| | Dense_layer_2 / Conv1D_layer_filter_2 | 200 | 128 |
| | Conv1D_layer_filter_3 | N/A | 256 |
| | kernel_size | N/A | 3 |
| | Dense_output / Conv1D_padding | 1 | same |
| **Adam Optimizer** | Learning Rate | 0.0003 | 0.0001 |
| | Betas | (0.5, 0.9) | |

For the Diffusion model, we employed the traditional U-Net architecture, replacing any 2D layers with their 1D counterparts. This included swapping Conv2D for Conv1D and Dropout2d for Dropout1d, allowing the U-Net to process our embedding vectors effectively. The structure of the network is detailed in Table 6.

Along with modifying the U-Net architecture, we made slight modifications to the Diffusion algorithm to better represent our 1D embedding vectors. Traditional Diffusion models use a time schedule to iteratively add and then remove noise from an image. At each time step, the image becomes a combination of more Gaussian noise and less of the original signal. When applying this traditional approach, we found that the Diffusion model struggled to accurately recreate our samples.

To address this issue, we modified the training approach to only add noise until timestep T/2 rather than the full T steps. This means that at the final timestep, the vector retains 96% of the Gaussian distribution, rather than 99%. This modification was applied symmetrically to both the forward and reverse diffusion processes. The revised reverse diffusion process is shown by Algorithm 1 (modified from [28] with T denoting the max timesteps and $\epsilon$ representing the normal distribution), enabled the model to produce adequate samples.

**TABLE 6.** Diffusion hyperparameters.

| | | Parameter | Details |
|---|---|-----------|---------|
| **Forward Diffusion** | | Linear Time Schedule | 1000 timesteps |
| **Reverse Diffusion** | | Layers | 4 encoding, 4 decoding |
| | | Base Channels | 32 |
| | | Channel Multipliers | (1, 2, 4, 8) |
| | U-Net Residual Block | Dropout | 0.01 |
| | | Time Embedding Multiplier | 2 |
| | | Applied Attention | true |
| | | Activation | Sigmoid Linear Unit (SiLU) |
| **AdamW Optimizer** | | Learning Rate | 5e-05 |
| | | Betas | (0.9, 0.999) |

---

**Algorithm 1** Sample Generation

**Input**: dataset $X_0$, num_samples
**Output**: synthetic samples
$X \leftarrow$ sampling($X_0$, num_samples)
$X_T \leftarrow$ forward_diffusion($X$, $T/2$) // replaced $\mathcal{N}(0, I)$
**for** $t = T, \ldots, 1$ **do**
    $\mathbf{z} \sim \mathcal{N}(\mathbf{0}, \mathbf{I})$ if $t > 1$, else $\mathbf{z} = 0$
    $\mathbf{x}_{t-1} = \frac{1}{\sqrt{\alpha_t}} \left( \mathbf{x}_t - \frac{1-\alpha_t}{\sqrt{1-\bar{\alpha}_t}} \epsilon_\theta(\mathbf{x}_t, t) \right) + \sigma_t \mathbf{z}$
**end**
return $\mathbf{x}_0$

---

## B. EXPERIMENTAL RESULTS

We evaluated our synthetic samples across three key categories: detection rate, high-quality data, and robustness. These were assessed using five metrics: binary classification, t-SNE visualization [29], thresholding for malware classification, multi-class classification, and similarity scores. For each classification metric, we utilized three different classifiers: Random Forest, Support Vector Machine (SVM), and Multilayer Perceptron (MLP). These classifiers were selected to capture a diverse range of learning approaches, ensuring robustness and completeness in our results.

### 1) HIGH QUALITY DATA

#### a: BINARY CLASSIFICATION

During Binary Classification, the model is trying to determine the difference between the synthetic and real samples. Our goal is to "fool" the model, meaning it is not able to correctly tell the difference between the two. Therefore, the ideal case would be an F1 score of 0.5, indicating the model cannot differentiate the synthetic samples from the real samples. However, in our experiments, we observe that the binary classifiers excel. By training on the synthetic samples, the classifiers can learn the intricate details that differentiate the synthetic and real samples, allowing them to achieve a high F1 score. Our analysis shows Renos, Enterak.A, and Beebone yield the best overall results, while Vobfus yields the worst results, as shown in Table 7. In terms of classifier performance, all three (Random Forest, SVM, and MLP) achieve comparable results, with MLP having a slightly higher F1 score. For both the best and worst-performing models, Diffusion and WGAN-GP consistently outperform GAN. In fact, Diffusion and WGAN-GP outperform GAN in 94% of the trials. Overall, these results show the strength of both the Diffusion and WGAN-GP models.

#### b: t-SNE VISUALIZATION

Because the malware embeddings, $1 \times 104$ vectors, are difficult to visualize directly, we utilized t-SNE as a dimension-reduction technique to transform these high-dimensional embeddings into 2D vectors. This approach enables us to visualize the malware embeddings and observe similarities between synthetic and real samples. Our hypothesis is that malware samples sharing similar features will be positioned close to each other in the reduced space. Consequently, distinct clusters of samples may correspond to different obfuscation techniques employed within the malware family.

**TABLE 7.** Binary classification family comparison.

| Family | Model | Random Forest | | SVM | | MLP | |
|---|---|---|---|---|---|---|---|
| | | Train | Test | Train | Test | Train | Test |
| Renos | Diffusion | 0.928 | 0.879 | 0.922 | 0.891 | 0.959 | 0.937 |
| | WGAN-GP | 0.920 | 0.849 | 0.877 | 0.857 | 0.948 | 0.935 |
| | GAN | 0.971 | 0.936 | 0.951 | 0.943 | 1.0 | 0.998 |
| Enterak.A | Diffusion | 0.970 | 0.915 | 0.803 | 0.789 | 0.992 | 0.983 |
| | WGAN-GP | 0.973 | 0.924 | 0.832 | 0.794 | 0.964 | 0.947 |
| | GAN | 0.988 | 0.953 | 0.902 | 0.892 | 1.0 | 1.0 |
| Beebone | Diffusion | 0.929 | 0.908 | 0.901 | 0.895 | 0.906 | 0.907 |
| | WGAN-GP | 0.961 | 0.949 | 0.926 | 0.919 | 0.936 | 0.934 |
| | GAN | 0.964 | 0.956 | 0.995 | 0.995 | 0.996 | 0.994 |
| Vobfus | Diffusion | 0.942 | 0.930 | 0.989 | 0.987 | 0.991 | 0.990 |
| | WGAN-GP | 0.912 | 0.903 | 0.993 | 0.991 | 0.992 | 0.986 |
| | GAN | 0.965 | 0.960 | 0.999 | 0.999 | 0.998 | 0.997 |

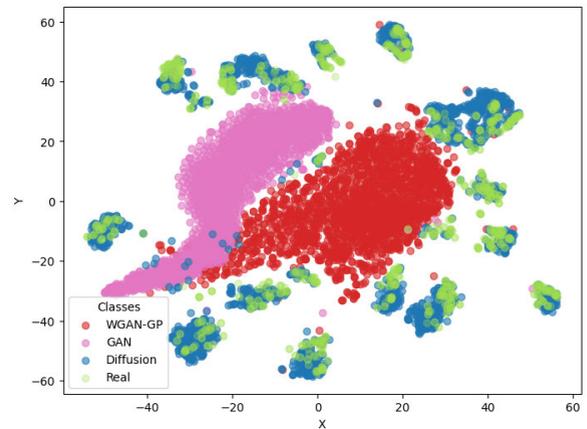

**FIGURE 6.** t-SNE dimension reduction on Expiro.BK. The visualization was generated using two components and a perplexity of 50.

Figure 6 depicts the t-SNE embeddings for the Expiro.BK malware family, which consists of 1,095 samples represented in green. These samples form a ring-like structure of distinct clusters. We choose to visualize this family because its unique distribution forces the models to represent a diverse range of malware techniques, unlike families with all their samples grouped in a single cluster.

When looking at the synthetic samples, the diffusion samples follow a very similar pattern to the baseline, showing they are similar to the real samples. In contrast, both the WGAN-GP and GAN samples form distinct, large groups in the middle of the ring. This highlights the disadvantages of the WGAN-GP and GAN models. While the diffusion synthetic samples are able to follow the patterns of the real data, the WGAN-GP and GAN models try to recreate all the samples at once. This results in a large group in the middle of the circle instead of distinct, smaller clusters like the original.

*c: SYNTHETIC-ONLY TRAINING EVALUATION*

We trained classifiers exclusively on synthetic malware and tested their performance on real malware data, adapting the concept of the Frechet Inception Distance (FID) metric, which is traditionally used to evaluate the similarity between generated and real two-dimensional data. Since our malware samples are one-dimensional, we extended this idea into two test cases: (1) training on synthetic data and testing on real data, and (2) training on real data and testing on synthetic data. This section focuses on the first case.

**TABLE 8.** Train Synthetic / Test Real: F1 Score accuracy.

| Model | Train Synthetic, Test Real | | |
|---|---|---|---|
| | RF | SVM | MLP |
| Diffusion | 0.88 | 1.00 | 0.99 |
| WGAN-GP | 0.89 | 1.00 | 0.99 |
| GAN | 0.73 | 0.96 | 0.96 |

The goal of this setup was to evaluate whether classifiers trained only on synthetic samples can effectively classify real-world malware. Table 10 shows the F1 Score of training solely on synthetic malware data, tested on real samples. Compared to the baseline F1 score of 0.87 (achieved from Table 1), the synthetic samples generated by Diffusion and WGAN-GP models demonstrated strong alignment with real data, achieving F1 scores close to the baseline. However, GAN-generated samples underperformed in this setup, with a significant 10% drop in the F1 score.

2) APPLICABILITY OF SYNTHETIC MALWARE

The previous experiment highlighted synthetic malware data's potential to enhance malware classification. However, it also emphasized the need for a tailored augmentation strategy to optimize classification performance. By introducing threshold constraints, this approach supplements real samples with high-quality synthetic data, effectively addressing the challenge of imbalanced datasets that often decrease the classification accuracy.

In this section, we introduce a threshold-based augmentation method designed to strategically supplement training data, ensuring effective augmentation that significantly improves classification accuracy. This approach empowers organizations to strengthen their malware classification capabilities, reduce the risk of undetected threats, and enhance the security of critical systems.

*a: THRESHOLD-BASED MALWARE CLASSIFICATION*

This method aims to optimize classification performance by determining the amount of synthetic data needed for each malware family based on a predefined threshold ($t$). Families with low baseline accuracy or with small amounts of samples were supplemented with synthetic data. Threshold values ranging from 0.90 to 0.99 were evaluated, and the best threshold ($t = 0.95$) was selected based on the highest

F1 score achieved by three generative models: Diffusion, WGAN-GP, and GAN.

Table 9 compares the baseline results (Table 4) with the augmented results under the threshold method (t = 0.95), both evaluated with the Random Forest classifier. The SVM and MLP classifiers both achieved an F1 score of 1.0 before applying the threshold method, indicating that under these classifiers, there is no room to gain any improvement. Thus, we excluded these 2 classifier's F1 scores from further analysis. Instead, we focused primarily on the Random Forest classifier, as it provided more informative and meaningful results. This method aims to show how carefully constrained augmentation can optimize malware classification performance. It demonstrates that the threshold-based augmentation significantly improves the F1 scores for all models, with Diffusion achieving the best performance, followed by WGAN-GP, and finally GAN.

**TABLE 9.** Threshold-based malware classification F1 Scores.

| Family | Model | Random Forest | |
|---|---|---|---|
| | | Baseline | Augmented |
| Overall | Diffusion | 0.87 | 0.951 |
| | WGAN-GP | 0.87 | 0.945 |
| | GAN | 0.87 | 0.942 |

Then, instead of comparing with baseline, we compare the results of augmentation before applying the threshold method (Figure 7-A, shown as the left bar chart) and after applying the threshold method (Figure 7-B, shown as the right bar chart). Both Figures show the difference in F1 score between the baseline model and the results of augmentations, with the middle black line representing no change between the two. Bars pointing to the right indicate improved F1 scores after augmentation, while those pointing to the left indicate decreases. Notably, after applying the threshold method, families with initially decreasing F1 scores showed significant improvement, particularly with the Diffusion and WGAN-GP models. Additionally, Togafn, a family that still experienced a slight decrease after applying threshold, nonetheless exhibited notable improvement compared to augmentation without thresholds.

*b: FIDELITY SCORE*

This method constitutes the second part of our experimental setup to replicate the Frechet Inception Distance (FID) metric, complementing the Synthetic-Only Training Evaluation. In this second part setup, classifiers were trained on real malware samples and tested exclusively on synthetic malware samples generated by each model. Higher F1 scores in this evaluation indicate a greater alignment of synthetic data with real malware, reflecting the models' ability to produce realistic and representative samples.

Table 10 presents the F1 scores achieved by different classifiers for each generative model in this experiment. Focusing on the Random Forest classifier, the synthetic malware generated by the Diffusion model achieved the highest F1 score at 92%, demonstrating the closest alignment with real data, followed by WGAN-GP at 90%. In contrast, the GAN model scored 10% lower than the other two models. These results underscore the fidelity of the synthetic samples within this evaluation setup.

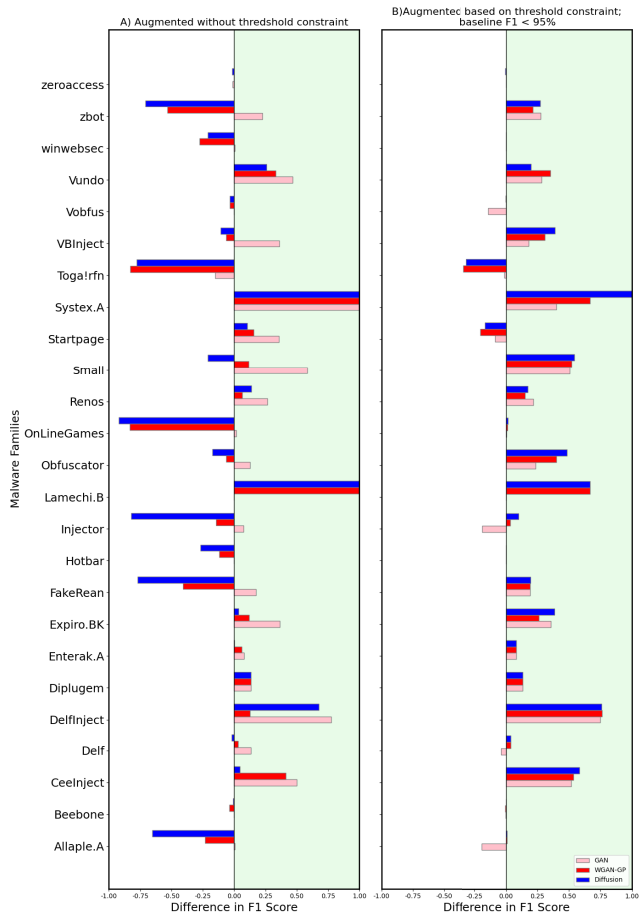

**FIGURE 7.** Differences in malware classification with data augmentation: The x-axis shows the change in F1 score after augmentation, with positive values indicating improvement and negative values indicating a decrease. The y-axis lists malware families. Chart (A) represents augmentation without the threshold method, while chart (B) applies the threshold (F1 < 0.95). This comparison demonstrates the effectiveness of threshold-based augmentation in enhancing classification performance.

**TABLE 10.** Train Real / Test Synthetic: F1 Score accuracy.

| Model | Train Real, Test Synthetic | | |
|-------|------|------|------|
| | RF | SVM | MLP |
| Diffusion | 0.92 | 1.00 | 1.00 |
| WGAN-GP | 0.90 | 1.00 | 1.00 |
| GAN | 0.81 | 0.92 | 0.95 |

### 3) ROBUSTNESS

To demonstrate robustness, we utilized cosine similarity by calculating a full cross-comparison between the real and synthetic samples for each generative model. A similarity score of 1 indicates that the vectors are very similar, whereas a minimum similarity score of $-1$ indicates the vectors are opposites. When computing the median similarity scores for each generative model, the goal is to achieve a similarity score identical to the cross-comparison of the baseline dataset itself. This shows that the distribution of the synthetic malware samples closely resembles the original.

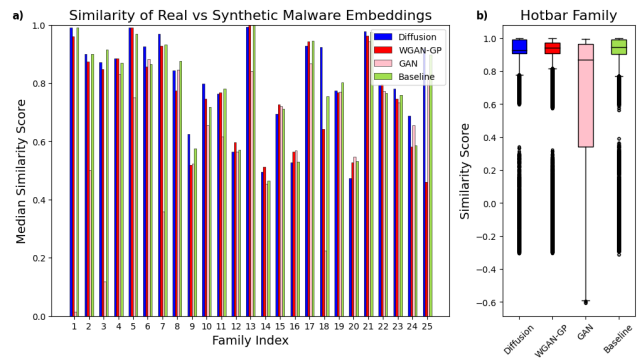

**FIGURE 8.** (a) The median similarity score for each malware family, based on the full cross-comparison between the real and synthetic datasets for each generative model. The baseline model compares the real dataset against itself. (b) Boxplot depicting the distribution of similarity scores between the real and synthetic datasets for the Hotbar family.

For most malware families, the median similarity scores across all four models are comparable. However, the GAN model clearly struggles, specifically for families 1, 3, 7, and 18, as shown in Figure 8a. In these cases, the GAN model achieves a significantly lower median similarity score compared to the other models. We hypothesize that this is due to overfitting on specific malware samples, leading to embeddings that are less representative of the overall distribution.

Moreover, these results highlight the robustness of the Diffusion and WGAN-GP models. On average, the Diffusion and WGAN-GP models achieve median similarity scores within 0.035 and 0.046 of the baseline, respectively. In contrast, the GAN model's synthetic samples are, on average, 0.176 away from the baseline.

Figure 8b further illustrates the GAN model's struggles using the Hotbar malware family. The boxplot for Diffusion and WGAN-GP closely match the baseline, with the first quartile deviating by just 0.004 and 0.006, respectively. Conversely, the GAN model's first quartile is 0.562 away from the baseline. Overall, these results signify the robustness of the Diffusion and WGAN-GP models, which consistently achieve results similar to the baseline without overfitting to any specific malware samples.

## V. DISCUSSION

This paper presents a system that leverages Generative AI techniques to create synthetic malware samples, addressing the challenge of imbalanced malware datasets. While GenAIs are effective at producing novel variations of known malware, they are inherently limited to the learned distribution and thus cannot generate truly unknown attack samples with fundamentally different characteristics. Instead, the generated samples closely resemble known malware families and serve to enrich training datasets by introducing diverse polymorphic variants. The primary objective remains to enhance malware classification, particularly when training data is limited. This section explores the strengths and limitations of the proposed system and outlines potential directions for future research.

### A. DATA DECOMPOSITION
The construction of training data is crucial for accurately representing malware attacks. To generate training data for our generative models, the proposed system decomposes malware binary samples into mnemonic opcode sequences. This approach focuses solely on opcodes, ignoring factors such as registers and addresses. The main advantage of this method is its simplicity, requiring fewer computational resources for the NLP model to learn and map malware embeddings to higher dimensions. However, omitting registers and addresses means the dataset does not fully capture the data flow in a malware attack. In the future, we plan to collect more comprehensive datasets by incorporating registers and addresses to enable a deeper analysis of data flow in synthetic malware data.

### B. SAMPLING EVALUATION TECHNIQUE
Before augmenting the training data, it is essential to assess the quality of the generated malware data thoroughly. The proposed system evaluates synthetic data using various metrics, including fidelity score, cosine similarity, and visual inspection of its proximity to real malware data. As discussed in the evaluation section, these metrics indicate that the synthetic data is of high quality, exhibits strong fidelity, and closely resembles authentic data. However, during data augmentation, all generated samples are selected rather than filtering out lower-quality ones. Implementing a sampling evaluation method would allow us to include only high-quality samples in the augmentation process, ultimately further improving malware classification performance.

In future work, we aim to develop a more advanced system for generating synthetic malware by addressing the challenges mentioned above. Our goal is to enhance robustness while exploring the potential of large language models (LLMs) in creating high-quality malware samples.

## VI. CONCLUSION
This paper proposes a new method for improving malware classification. Utilizing three generative models - Diffusion, WGAN-GP, and GAN - modified to handle 1-D data, we create synthetic malware samples that can be used to supplement training data. Our samples are evaluated using six key metrics: binary classification, t-SNE visualization, synthetic-only training evaluation, application of synthetic malware, fidelity score, and robustness. Significant improvements were shown in the application of synthetic malware. By individually supplementing the training set for specific malware families, we are able to increase the overall F1 Score for multiclass classification from 87% to 95% with diffusion samples. Our other metrics reinforce the high quality of our samples, showing a high fidelity score, robustness, and a large diversity in generated samples. Ultimately, these experimental results signify the strength of the Diffusion model and the positive impact of utilizing synthetic malware samples to improve malware classification.

In future work, this research could follow a variety of directions. Various natural language processing techniques present different strengths when compared to Word2Vec, such as ELMo, GloVe, and BERT [7]. Additionally, exploring alternative generative models could yield improvements. Specifically, diffusion models focused on naturally discrete data show promise; namely discrete diffusion and embedding diffusion models may provide valuable insights and create more promising opportunities for addressing the challenges of insufficient malware data.

## ACKNOWLEDGMENT
*(Tiffany Bao, Kylie Trousil, and Quang Duy Tran contributed equally to this work.)*

The authors would like to express their deepest gratitude to Dr. Younghee Park and Dr. Fabio Di Troia for their invaluable guidance and mentorship throughout this research project. They also extend their thanks to the graduate students who provided insightful support and guidance during the summer research program. Special thanks to San José State University for serving as the REU site and for providing the resources and collaborative environment that made this work possible.

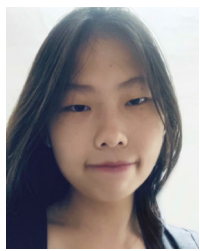

**TIFFANY BAO** is currently pursuing the bachelor's degree in mathematics and computer science with Boston University. Her research interests include machine learning, cybersecurity, and information security.

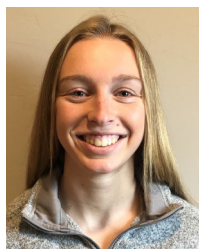

**KYLIE TROUSIL** is currently pursuing the bachelor's degree in applied mathematics and computer science with the University of Wisconsin–La Crosse. Her research interests include machine learning, artificial intelligence, and cybersecurity.

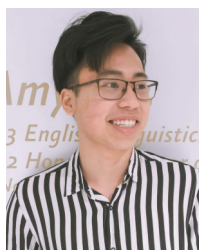

**QUANG DUY TRAN** is currently pursuing the master's degree in data science with San José State University. He is a Teaching Associate with the Department of Computer Science, University's. He plans to pursue the Ph.D. degree in computer science in USA, focusing on research in cybersecurity, including machine learning, malware/intrusion detection, and AI security.

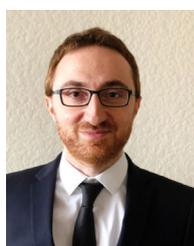

**FABIO DI TROIA** (Member, IEEE) received the Ph.D. degree in computer science from Kingston University, London, U.K., researching applications of machine learning in the field of cybersecurity. He is currently an Assistant Professor with the Department of Computer Science, San José State University, teaching information security and machine learning bachelor's and graduate courses. In particular, his areas of focus are malware detection, malware design, cryptography, network analysis, and access control. During his career, he supervised more than 200 bachelor's and graduate students, accomplishing publications in major journals and conference proceedings. Since 2020, he has been part of the Silicon Valley Cybersecurity Institute (SVCSI), a nonprofit organization that engages local communities in diverse cybersecurity activities to support the development of the future cybersecurity workforce.

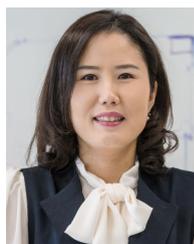

**YOUNGHEE PARK** (Member, IEEE) received the Ph.D. degree in computer science from North Carolina State University. She was a Visiting Professor with IBM Almaden Research, in 2019. She is currently an Associate Professor of computer engineering with San José State University. She is the President and the Founder of the Silicon Valley Cybersecurity Institute. Her research projects have been funded by NSF, DoE, Google, and Cisco. Her research interests include cybersecurity and education, including machine learning, SDN/NFV security, biometric security, blockchain security, and the IoT security. She received the SJSU Distinguished Faculty Mentor Award, in 2017, and the Faculty Award for Excellence in Scholarship, in 2018. She was the Kordestani Endowed Chair at the College of Engineering Research Professor Award, from 2016 to 2017. She received the Best Paper Award at ACM SIGCSE 2018 and the Best Paper Honorable Award at ACM SACMAT 2016.

• • •